\documentclass[10pt,twocolumn,letterpaper]{article}

\usepackage{cvpr}
\usepackage{caption}

\usepackage[pagebackref,breaklinks,colorlinks,allcolors=cvprblue]{hyperref}
\usepackage[capitalize] {cleveref}
\usepackage{algorithm}
\usepackage{algorithmic}
\usepackage{booktabs}
\usepackage{cleveref}
\usepackage{tcolorbox}
\tcbuselibrary{skins}
\usepackage{tabularx,booktabs,ragged2e}

\definecolor{cvprblue}{rgb}{0.21,0.49,0.74}
\usepackage[pagebackref,breaklinks,colorlinks,allcolors=cvprblue]{hyperref}
\usepackage{subcaption}   
\usepackage{adjustbox}    
\usepackage{xcolor}
\definecolor{darkgreen}{rgb}{0.0, 0.5, 0.0}
\def\paperID{7} 
\def\confName{CVPR}
\def\confYear{2026}

\title{Iterative Definition Refinement for Zero-Shot Classification via LLM-Based Semantic Prototype Optimization}

\author{Naeem Rehmat \and Muhammad Saad Saeed \and Ijaz Ul Haq \and Khalid Malik\\
University of Michigan-Flint, Michigan, USA\\
{\tt\small \{nrehmat, msaads, ijazhaq, drmalik\}@umich.edu
}}

\begin{document}
\maketitle

\begin{abstract}

Web filtering systems rely on accurate web content classification to block cyber threats, prevent data exfiltration, and ensure compliance. However, classification is increasingly difficult due to the dynamic and rapidly evolving nature of the modern web. Embedding-based zero-shot approaches map content and category descriptions into a shared semantic space, enabling label assignment without labeled training data, but remain highly sensitive to definition quality. Poorly specified or ambiguous definitions create semantic overlap in the embedding space, leading to systematic misclassification.

In this paper, we propose a training-free, adaptive iterative definition refinement framework that improves zero-shot web content classification by progressively optimizing category definitions rather than updating model parameters. Using LLMs as feedback-driven definition optimizers, we investigate three refinement strategies namely example-guided, confusion-aware, and history-aware, each refining class descriptions using structured signals from misclassified instances. Furthermore, we introduce a human-labeled benchmark of 10 URL categories with 1,000 samples per class and evaluate across 13 state-of-the-art embedding foundation models. Results demonstrate that iterative definition refinement consistently improves classification performance across diverse architectures, establishing definition quality as a critical and underexplored factor in embedding-based systems. The dataset is available \href{https://github.com/naeemrehmat/B2MWT-10C}{here}.

\end{abstract}

\section{Introduction}
\label{sec:intro}

\begin{figure}[t]
    \centering
    \includegraphics[width=1.0\linewidth]{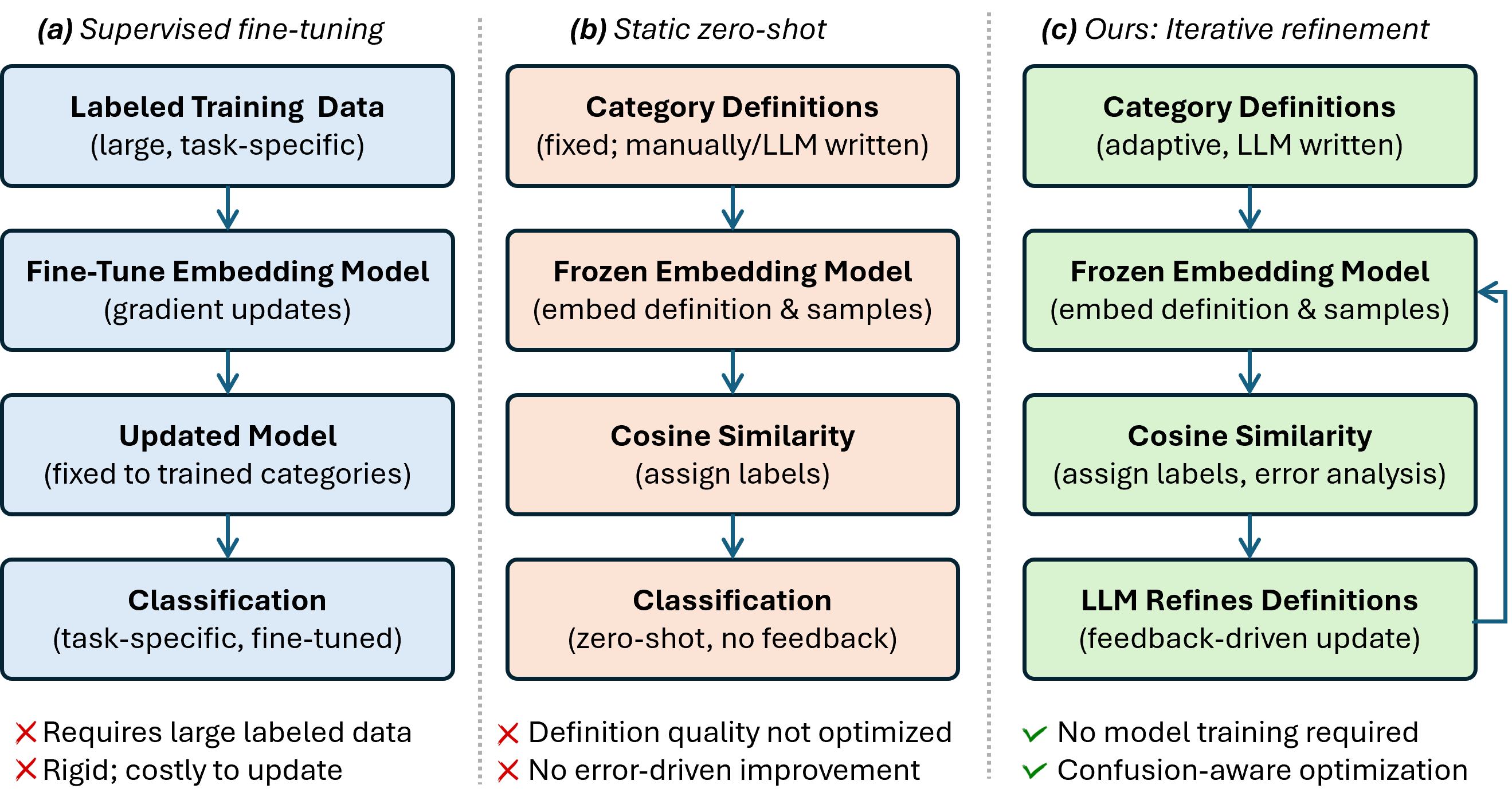}
    \caption{Comparison of URLs classification paradigms. Unlike (a) supervised fine-tuning and (b) static zero-shot methods, our proposed approach (c) iteratively refines category definitions using LLM feedback, without updating model parameters.}
    \label{fig:motivation}
\end{figure}

The classification of web contents into semantic categories is a critical component in a wide range of applications, including content filtering, malicious URL detection, network traffic analysis, and web indexing \cite{le2018urlnet, saxe2017expose}. Traditional supervised approaches to URL classification require large annotated datasets and
are brittle to the rapidly evolving nature of web content ~\cite{maneriker2021urltran}. Embedding-based zero-shot classification offers an alternative by representing both URLs and category descriptions in a shared semantic space, enabling
label inference through similarity computation without task-specific training ~\cite{reimers2019sentence, yin2019benchmarking}. The effectiveness of this paradigm, however, is fundamentally tied to the quality of the category definitions. In embedding space, each category definition acts as a semantic prototype where the class boundary is implicitly determined by the position of this prototype relative to document embeddings \cite{wang2022text, muennighoff2023mteb}. When definitions are vague, overly general, or semantically similar across classes, the resulting prototype placement causes systematic misclassifications even when the underlying embedding model possesses strong representational capacity. 

As illustrated in Figure~\ref{fig:motivation}, this limitation is shared by both supervised fine-tuning approaches ~\cite{yin2019benchmarking} which update model parameters at significant cost and static zero-shot methods ~\cite{su2023one}. This observation motivates a key question: can classification performance be improved by refining the category definitions themselves, rather than by retraining or fine-tuning the embedding model? We answer this question affirmatively by proposing an iterative definition refinement framework in which LLMs are used to progressively improve class definitions based on feedback from classification results. The proposed framework introduces three complementary refinement strategies.
Example-guided refinement updates definitions using representative samples to ground them in real data.
Confusion-aware refinement targets the most frequently misclassified class pairs to improve discriminability at decision boundaries.History-aware refinement incorporates the trajectory of previous definitions and performance scores to guide further improvements, enabling a form of iterative knowledge evolution. Our main contributions are as follows:
\begin{enumerate}[label=(\arabic*)]
 
\item We introduce the first framework that applies
iterative semantic prototype refinement to web content classification, achieving competitive performance without model training or fine-tuning.
Unlike prior supervised~\cite{maneriker2021urltran, le2018urlnet} and static zero-shot approaches ~\cite{yin2019benchmarking, su2023one}, our framework dynamically adapts category definitions from classification feedback, leaving all embedding models frozen.
  
 \item We propose \textit{example-guided}, \textit{confusion-aware}, and \textit{history-aware} refinement each targeting a distinct failure mode (data grounding, inter-class confusion, and definition trajectory). specifically we combined Monte Carlo acceptance rule in history-aware strategy, enabling principled exploration of the definition space without getting trapped in local optima.

 \item We systematically benchmark all strategy-model combinations (13 state-of-the-art embedding foundation models and 4 LLM), identifying which pairings yield the most discriminative definitions and providing deployment-ready guidance for practitioners.
  
  \item We release a B2MWT-10C (Benchmark Multilingual Multi-Web Text)  dataset of 10,000 URLs across 10 semantically overlapping categories.

  \end{enumerate}

\section{Related Work}
\label{sec:Related_work}
Early web content classification methods relied on hand-crafted lexical and structural features extracted from URL strings, including token n-grams, character-frequency distributions, and domain-level statistics. Deep learning shifted the paradigm toward end-to-end representation learning. Convolutional neural networks (CNNs) applied at the character level demonstrated strong performance for malicious URL detection, learning local structural patterns indicative of phishing and spam \cite{le2018urlnet, saxe2017expose}. Recurrent architectures, particularly LSTMs, subsequently captured
sequential dependencies across URL path and query components \cite{huang2020malicious}.
Transformer-based models fine-tuned on URL classification tasks achieved state-of-the-art results on standard benchmarks \cite{devlin2019bert}, benefiting from subword tokenization and contextual pre-training. More recently, graph neural networks have incorporated link structure and co-occurrence patterns, framing URL classification as node classification in a web graph \cite{yang2021webphish}. Despite their strong supervised performance, these approaches require substantial labeled data and retraining whenever category definitions evolve, and offer no mechanism for zero-shot generalization to novel categories.

Zero-shot classification methods address the labeled-data bottleneck. Early approaches based on natural language inference (NLI) frame classification as textual entailment, scoring candidate labels via a premise–hypothesis structure \cite{yin2019benchmarking}. Embedding-based classification represents documents and category definitions independently in a shared vector space and assigns labels via cosine similarity \cite{reimers2019sentence, pratt2023does}. The quality of embeddings from modern sentence transformers and instruction-tuned models \cite{wang2022text, su2023one, Xiao2024BGE} has made this paradigm increasingly competitive, with comprehensive benchmarking provided by MTEB \cite{muennighoff2023mteb}. LLMs have also been applied directly to classification through prompt engineering \cite{brown2020language, touvron2023llama}, where the model assigns a label given the input and a description of candidate categories. While effective, this requires a full LLM forward pass per sample, which is computationally prohibitive at scale. Automatic prompt optimization methods \cite{zhou2022large, pryzant2023automatic} are conceptually related. These methods improve LLM performance by modifying inputs rather than weights. However, these methods target generative tasks and optimize prompts for auto regressive models.
Our framework instead optimizes semantic prototype definitions for embedding-based classifiers, involving different objectives, feedback signals, and optimization dynamics.
To our knowledge, no prior work has systematically studied iterative definition refinement for zero-shot web content classification across diverse embedding architectures.


\begin{figure*}[!t]
    \centering
    \includegraphics[width=\linewidth]{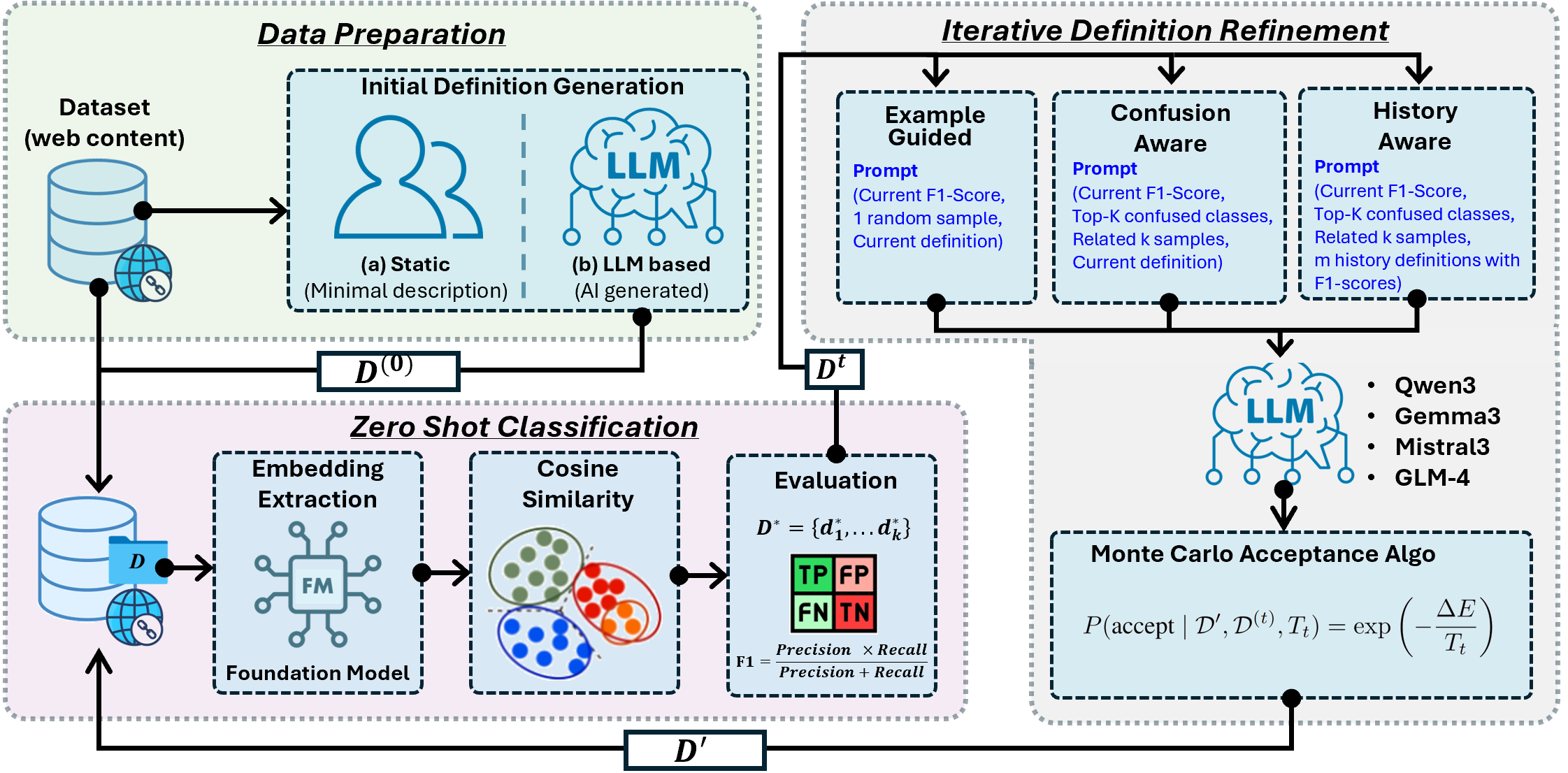}
    \caption{Workflow diagram of proposed method.}
    \label{fig:workflow}
\end{figure*}

\section{Methodology}
\label{sec:method}

\subsection{Problem Formulation}

Let $\mathcal{X} = \{(x_i, y_i)\}_{i=1}^{n}$ denote a labeled dataset where $x_i$ represents the textual content extracted from the HTML of a webpage associated with a URL and $y_i \in \mathcal{C} = \{c_1, \ldots, c_k\}$ denotes the ground truth category. For each class $c_j$, we maintain a \emph{textual definition} $d_j$; a short natural-language description of the category; that serves as a semantic prototype in embedding space. We assume a fixed embedding function $f$ that maps any text to a $d$-dimensional vector: $f : \text{text} \rightarrow \mathbb{R}^d$ (e.g., from a pretrained sentence encoder). The embedding of a definition, $f(d_j)$, is the vector representation of the definition of class $c_j$; the classifier assigns to a webpage $x$ the class whose definition embedding is most similar to $f(x)$. Using cosine similarity, the prediction rule is:

\begin{equation}
\hat{y} = \arg\max_{j \in \{1,\ldots,k\}} \cos\big(f(x), f(d_j)\big)
\end{equation}

The goal of our framework is to find a set of definitions

\begin{equation}
   \mathcal{D}^* = \{d_1^*, \ldots, d_k^*\}
\end{equation}

that maximizes classification performance over $\mathcal{X}$, measured by \emph{Macro F1}, without updating the embedding function $f$. Figure~\ref{fig:workflow} shows the workflow of proposed framework.

\subsection{Zero-Shot Classification Framework}
\label{sec:framework}
The classification pipeline operates as follows.
\begin{enumerate}
    \item Category definitions can be initialized using two approaches: \
\textbf{Minimal:} each category is described as ``A webpage about \{category\}.'' \\ \textbf{LLM-generated:} an LLM is prompted to provide a concise definition of the category {category}.''; we adopt this approach, and these definitions form the initial knowledge base for future refinement.
    

    \item HTML text and category definitions are embedded using a pretrained embedding model. Some embedding models require task-specific instruction prefixes so that embeddings are generated in the intended configuration. For example, \texttt{multilingual-e5-large} uses the prefix ``query:'', \texttt{llama-nemotron-embed-1b-v2} uses ``passage:'', and \texttt{modernbert-embed-large} uses ``classification:''. Other models such as \texttt{bge-m3} and \texttt{mxbai-embed-large-v1} do not require instruction prefixes.
    \item Cosine similarity between each HTML embedding and each definition embedding is computed.
    \item The label assigned is the category with the highest similarity score.
\end{enumerate}

In this setup, each category definition acts as a semantic prototype in the embedding space; classification quality therefore depends on how well these prototypes separate the classes.

\begin{figure*}[t]
    \centering
    \includegraphics[width=1.0\linewidth]{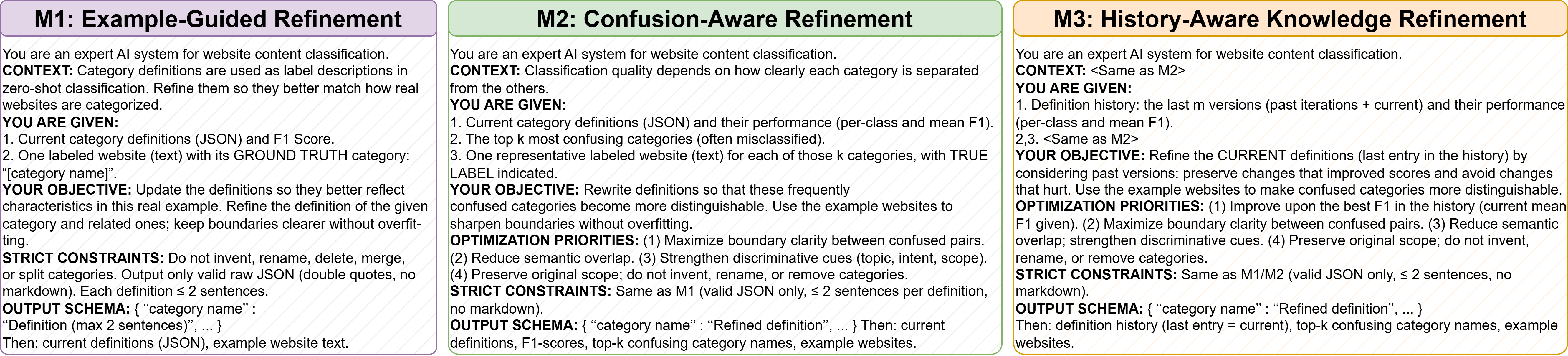}
    \caption{Prompt templates used for the three refinement strategies.}
    \label{fig:prompt_templates}
\end{figure*}

\subsection{Iterative Definition Refinement}
\label{sec:refinement}

We iteratively refine category definitions using feedback from classification performance. At each iteration, the current definitions are provided to an LLM together with task-specific feedback derived from predictions of the zero-shot classification framework described in \S\ref{sec:framework}. Based on this feedback, the LLM generates revised definitions intended to better separate categories in embedding space. The updated definitions are then used to re-run the classifier, producing a new Macro F1 score that serves as feedback for the next iteration.

The prompt templates used for these refinement strategies are illustrated in Figure~\ref{fig:prompt_templates}. All refinement strategies share this iterative loop but differ in the evidence provided to the LLM. In all cases, the LLM receives the current definitions and the Macro F1 score obtained under them. Additional feedback may include labeled webpage instances, samples from frequently confused class pairs, or a history of previous definitions and their scores. We describe three refinement strategies of increasing sophistication.

\subsubsection{M1: Example-Guided Refinement}

In this strategy, the LLM receives the current definitions, a randomly sampled labeled webpage instance (the textual content together with its ground truth label), and the Macro F1 score obtained under the current definitions. The LLM is prompted to revise the definitions so that they better capture the semantic characteristics of instances belonging to each category.

\subsubsection{M2: Confusion-Aware Refinement}

This strategy augments the feedback with examples drawn from frequently confused categories. The LLM receives the current definitions, one random labeled webpage content sampled from the class pairs with the highest $k$ misclassified categories from the confusion matrix, and the Macro F1 score under the current definitions. These examples highlight decision boundaries where the classifier performs poorly. The LLM is prompted to revise the definitions to make frequently confused categories more distinguishable in embedding space. We vary $k \in \{1,2,3,4,5\}$.

\subsubsection{M3: History-Aware Refinement with Stochastic Acceptance}
\label{sec:history-boltzmann}

The history-aware strategy further extends confusion-aware refinement by providing the LLM with the $m$ most recent sets of definitions together with their corresponding Macro F1 scores. This allows the LLM to reason about the trajectory of definition evolution, identifying which changes improved performance and which degraded it. Hyperparameters $k \in \{1,2,3,4,5\}$ and $m \in \{1,2,3,4,5\}$ control the size of the feedback window.

To avoid getting trapped in local optima, we adopt a \emph{Monte Carlo} acceptance rule inspired by the Metropolis criterion~\cite{metropolis1953equation} and simulated annealing~\cite{kirkpatrick1983optimization}. Let $\mathcal{D}^{(t)}$ denote the current set of definitions at iteration $t$, and $\mathcal{D}'$ a candidate set proposed by the LLM. Let $\phi(\mathcal{D})$ denote the Macro F1 score of the classifier when using definitions $\mathcal{D}$. We define the ``energy'' drop when moving from the current to the proposed state as
\begin{equation}
\Delta E = \phi(\mathcal{D}_{\text{best}}) - \phi(\mathcal{D}').
\label{eq:energy-drop}
\end{equation}
If the proposed definitions improve performance ($\Delta E \leq 0$), we always accept:
\begin{equation}
\mathcal{D}^{(t+1)} = \mathcal{D}' \quad \text{if } \phi(\mathcal{D}') \geq \phi(\mathcal{D}_{\text{best}}).
\label{eq:accept-improvement}
\end{equation}
Otherwise, we accept the worse candidate with probability given by the \emph{Boltzmann acceptance rule}:
\begin{equation}
P(\text{accept} \mid \mathcal{D}', \mathcal{D}_{\text{best}}, T_t) = \exp\left( -\frac{\Delta E}{T_t} \right),
\label{eq:boltzmann}
\end{equation}
where $T_t > 0$ is a temperature parameter at iteration $t$. Thus $\mathcal{D}^{(t+1)} = \mathcal{D}'$ with probability~\eqref{eq:boltzmann}, and $\mathcal{D}^{(t+1)} = \mathcal{D}^{(t)}$ otherwise. We use a simple annealing schedule so that exploration is reduced over time:
\begin{equation}
T_t = \max\left( T_{\min},\; T_0 \left(1 - \frac{t}{T_{\max}}\right) \right),
\label{eq:temperature}
\end{equation}
with $T_0$, $T_{\min}$, and $T_{\max}$ fixed (e.g., $T_0 = 0.1$, $T_{\min} = 0.01$, $T_{\max}$ = maximum iterations). This allows occasional acceptance of worse definitions early in the process while converging toward greedy acceptance as $t \to T_{\max}$~\cite{kirkpatrick1983optimization}. Algorithm~\ref{alg:refine} summarizes the complete History-Aware Refinement with Stochastic Acceptance procedure.

\begin{algorithm}[t]
\footnotesize
\caption{History-Aware Refinement with Simulated Annealing}
\label{alg:refine}
\begin{algorithmic}[1]
\REQUIRE Labeled dataset $\mathcal{X}_{\text{train}}$; initial definitions $\mathcal{D}^{(0)}$;
         embedding model $f$ (frozen); max iterations $T_{\max}$;
         initial temperature $T_0$; hyperparameters $k$, $m$.
\ENSURE Refined definitions $\mathcal{D}_{\text{best}}$ and best Macro F1 $\phi_{\text{best}}$.

\STATE $\mathcal{D} \gets \mathcal{D}^{(0)}$;
       $\phi_{\text{cur}} \gets \phi_{\text{train}}(\mathcal{D})$;
       $\phi_{\text{best}} \gets \phi_{\text{cur}}$;
       $\mathcal{D}_{\text{best}} \gets \mathcal{D}$
\STATE $\mathcal{H} \gets \bigl[(\mathcal{D},\; \phi_{\text{cur}})\bigr]$
       \COMMENT{History of last $m$ accepted definition sets}

\FOR{$t = 1$ \TO $T_{\max}$}
    \STATE Compute top-$k$ confused class pairs from train confusion matrix under $\mathcal{D}$;
           sample one train article per confused class
    \STATE $\mathcal{D}' \gets \textsc{ProposeDefinitions}(\mathcal{D},\; \phi_{\text{cur}},\; \mathcal{H},\;
           \text{top-}k\text{ pairs},\; \text{samples})$
           \COMMENT{LLM call with history and confusion context}
    \STATE $\phi' \gets \phi_{\text{train}}(\mathcal{D}')$
    \STATE $\Delta E \gets \phi_{\text{best}} - \phi'$;
           $T_t \gets \max\bigl(T_{\min},\; T_0 \cdot (1 - t / T_{\max})\bigr)$;

    \IF{$\phi' \geq \phi_{\text{best}}$}
        \STATE $\mathcal{D} \gets \mathcal{D}'$;
               $\phi_{\text{cur}} \gets \phi'$;
               $\mathit{accepted} \gets \mathrm{True}$
               \COMMENT{Always accept improvement over best}
    \ELSE
        \STATE Sample $u \sim \mathrm{Uniform}(0,1)$
        \IF{$u < \exp(-\Delta E / T_t)$}
            \STATE $\mathcal{D} \gets \mathcal{D}'$;
                   $\phi_{\text{cur}} \gets \phi'$;
                   $\mathit{accepted} \gets \mathrm{True}$
                   \COMMENT{Boltzmann acceptance of downhill move}
        \ENDIF
    \ENDIF

    \IF{$\phi_{\text{cur}} > \phi_{\text{best}}$}
        \STATE $\phi_{\text{best}} \gets \phi_{\text{cur}}$;
               $\mathcal{D}_{\text{best}} \gets \mathcal{D}$
    \ENDIF
    \IF{$\mathit{accepted}$}
        \STATE Append $(\mathcal{D},\; \phi_{\text{cur}})$ to $\mathcal{H}$;
               retain only last $m$ entries
               \COMMENT{Rejected proposals excluded from history}
    \ENDIF
\ENDFOR
\STATE \RETURN $\mathcal{D}_{\text{best}}$, $\phi_{\text{best}}$
\end{algorithmic}
\end{algorithm}

\section{Experimental Setup}
\label{sec:experiments}


\begin{table}[b]
\centering
\scriptsize
\setlength{\tabcolsep}{2pt}
\begin{subtable}[c]{0.49\columnwidth}
\centering
\begin{adjustbox}{max width=\linewidth}
\begin{tabular}{lccc}
\hline
Category & Train & Dev & Test \\
\hline
Art \& Design & 200 & 257 & 313 \\
Books & 200 & 284 & 305 \\
Dance & 200 & 318 & 318 \\
Fashion \& Style & 200 & 309 & 307 \\
Food & 200 & 326 & 290 \\
Health & 200 & 285 & 307 \\
Media & 200 & 312 & 326 \\
Movies & 200 & 325 & 292 \\
Music & 200 & 299 & 286 \\
Opinion & 200 & 280 & 284 \\
Real Estate & 200 & 289 & 299 \\
Science & 200 & 308 & 306 \\
Sports & 200 & 314 & 280 \\
Technology & 200 & 300 & 325 \\
Television & 200 & 301 & 281 \\
Theater & 200 & 281 & 311 \\
Travel & 200 & 301 & 287 \\
\hline
\end{tabular}
\end{adjustbox}
\caption{N24News}
\label{tab:n24_dataset}
\end{subtable}\hfill
\begin{subtable}[c]{0.49\columnwidth}
\centering
\begin{adjustbox}{max width=\linewidth}
\begin{tabular}{lccc}
\hline
Category & Train & Dev & Test \\
\hline
Betting \& Gambling & 200 & 200 & 600 \\
Business \& Industry & 200 & 200 & 600 \\
Content Retrieval Issues & 200 & 200 & 600 \\
Cultural \& Creative Arts & 200 & 200 & 600 \\
Leisure \& Entertainment & 200 & 200 & 600 \\
Literature \& Books & 200 & 200 & 600 \\
Pornographic Content & 200 & 200 & 600 \\
Retail eCommerce & 200 & 200 & 600 \\
Unused \& Placeholder & 200 & 200 & 600 \\
Video \& Online Games & 200 & 200 & 600 \\
\hline
\end{tabular}
\end{adjustbox}
\caption{B2MWT-10C (Ours)}
\label{tab:b2mwt_dataset}
\end{subtable}
\caption{Distribution of sample counts across categories in the datasets used in our experiments.}
\label{tab:dataset_stats}
\end{table}

\subsection{Datasets}
\label{sec:dataset}

We evaluate our methods on two datasets: the publicly available \textit{N24News} benchmark~\cite{wang2022n24news} and a newly constructed URL classification dataset. Table~\ref{tab:dataset_stats} summarizes the category distributions used in our experiments.

\textbf{N24News Dataset.} The N24News dataset~\cite{wang2022n24news} is a large-scale news classification benchmark consisting of news articles categorized into multiple topical classes. For our experiments, we select categories containing exactly 3,000 samples, resulting in a subset of 17 categories. From the training portion of each category, we randomly sample 200 instances to support definition refinement during the iterative process. Whereas the number of dev and test samples per category is shown in Table~\ref{tab:n24_dataset}. Dev set is used to track performance improvement and test set to report final results.

\textbf{B2MWT-10C Dataset.} We introduce a new human-labeled benchmark for content-based URL classification. The dataset contains 10 categories (10C) with 1,000 samples per class  as give in Table~\ref{tab:b2mwt_dataset}. All samples are assigned a single category by human annotators to ensure high-quality ground truth. For final evaluation, we reserve 600 samples per class as the test set, 200 for dev set which is used for tracking performance improvement, and use the remaining 200 as train set for the refinement process. 

\subsection{Embedding Models}

We evaluate thirteen text embedding models, all with fewer than 1B parameters, spanning
diverse architectures and training objectives. The models can be broadly categorized into
three groups: general-purpose multilingual models, including Alibaba-NLP/gte
\cite{zhang2024mgte} (305M), BAAI/bge-m3 \cite{chen2024bge} (567M),
Multilingual-e5-large \cite{wang2024multilingual} (560M), Snowflake-arctic-embed-v2.0
\cite{yu2024arcticembed} (568M), and Granite-embedding-278m \cite{awasthy2025granite}
(278M); LLM-derived embedding models, including KaLM-Embedding-v2.5
\cite{hu2025kalmembedding} (494M), Qwen3-Embedding-0.6B
\cite{zhang2025qwen3embedding} (600M), Embeddinggemma-300m
\cite{schechter2025embeddinggemma} (300M), Llama-nemotron-embed-1b-v2
\cite{nemotron2025embed1b} (1B), and Voyage-4-nano \cite{voyage2026nano} (340M); and
encoder-based models, including ModernBERT-embed-large \cite{warner2025modernbert}
(395M), Mxbai-embed-large-v1 \cite{lee2024mxbai} (335M), and
Solon-embeddings-large-0.1 \cite{solon2024} (560M).

\begin{table*}[t]
\centering
\small
\definecolor{darkgreen}{rgb}{0.0, 0.5, 0.0}
\resizebox{\textwidth}{!}{%
\begin{tabular}{l|c|cc|cc|cc}
\hline
\textbf{Models} & \textbf{B} & \textbf{M1 (LLM)} & \textbf{M1} & \textbf{M2 (LLM + k)} & \textbf{M2} & \textbf{M3 (LLM + k,m)} & \textbf{M3} \\
\hline
Alibaba-NLP\_gte & 52.14 & mistral & 59.16{\color{darkgreen}$^{+7.02}$} & qwen (k=5) & 59.25{\color{darkgreen}$^{+7.11}$} & mistral (k=5,m=1) & 60.44{\color{darkgreen}$^{+8.30}$} \\
BAAI\_bge-m3 & 52.75 & gemma & 59.67{\color{darkgreen}$^{+6.92}$} & mistral (k=2) & 59.82{\color{darkgreen}$^{+7.07}$} & mistral (k=5,m=1) & 60.60{\color{darkgreen}$^{+7.85}$} \\
KaLM-Embedding-v2.5 & 59.60 & mistral & 68.62{\color{darkgreen}$^{+9.02}$} & mistral (k=1) & 69.87{\color{darkgreen}$^{+10.27}$} & mistral (k=5,m=1) & 68.52{\color{darkgreen}$^{+8.92}$} \\
Solon-embeddings-large-0.1 & 48.12 & mistral & 58.01{\color{darkgreen}$^{+9.89}$} & mistral (k=1) & 57.92{\color{darkgreen}$^{+9.80}$} & mistral (k=2,m=1) & 58.38{\color{darkgreen}$^{+10.26}$} \\
Qwen3-Embedding-0.6B & 56.15 & glm-4 & 68.17{\color{darkgreen}$^{+12.02}$} & mistral (k=5) & 68.06{\color{darkgreen}$^{+11.91}$} & mistral (k=5,m=1) & 67.54{\color{darkgreen}$^{+11.39}$} \\
Snowflake-arctic-embed-v2.0 & 50.10 & qwen & 61.37{\color{darkgreen}$^{+11.27}$} & mistral (k=2) & 59.59{\color{darkgreen}$^{+9.49}$} & mistral (k=5,m=1) & 61.79{\color{darkgreen}$^{+11.69}$} \\
Embeddinggemma-300m & 48.64 & qwen & 66.39{\color{darkgreen}$^{+17.75}$} & gemma (k=4) & 65.82{\color{darkgreen}$^{+17.18}$} & mistral (k=5,m=1) & 67.65{\color{darkgreen}$^{+19.01}$} \\
Granite-278m-multilingual & 52.18 & qwen & 57.55{\color{darkgreen}$^{+5.37}$} & qwen (k=2) & 55.46{\color{darkgreen}$^{+3.28}$} & mistral (k=4,m=5) & 57.00{\color{darkgreen}$^{+4.82}$} \\
Multilingual-e5-large & 41.77 & mistral & 46.48{\color{darkgreen}$^{+4.71}$} & mistral (k=3) & 47.34{\color{darkgreen}$^{+5.57}$} & mistral (k=4,m=4) & 49.07{\color{darkgreen}$^{+7.30}$} \\
Modernbert-embed-large & 55.02 & qwen & 65.69{\color{darkgreen}$^{+10.67}$} & mistral (k=3) & 62.30{\color{darkgreen}$^{+7.28}$} & glm-4 (k=3,m=2) & 64.65{\color{darkgreen}$^{+9.63}$} \\
Mxbai-embed-large-v1 & 54.75 & glm-4 & 61.93{\color{darkgreen}$^{+7.18}$} & mistral (k=2) & 61.06{\color{darkgreen}$^{+6.31}$} & mistral (k=5,m=1) & 63.50{\color{darkgreen}$^{+8.75}$} \\
Llama-nemotron-embed-1b-v2 & 49.19 & qwen & 62.68{\color{darkgreen}$^{+13.49}$} & gemma (k=4) & 60.29{\color{darkgreen}$^{+11.10}$} & gemma (k=3,m=1) & 60.82{\color{darkgreen}$^{+11.63}$} \\
Voyage-4-nano & 62.65 & mistral & 74.80{\color{darkgreen}$^{+12.15}$} & mistral (k=3) & 74.36{\color{darkgreen}$^{+11.71}$} & mistral (k=4,m=4) & \textbf{75.49}{\color{darkgreen}$^{+12.84}$} \\
\hline
\end{tabular}%
}
\caption{F1 scores of embedding models on the B2MWT-10C test set. We compare the baseline (B) with methods M1, M2, and M3. Superscripts denote the absolute improvement over the baseline ($\Delta = \text{M} - \text{B}$), and \textcolor{darkgreen}{green} indicates a positive gain.}
\label{tab:netstar_results}
\end{table*}

\begin{table*}[h!]
\centering
\small
\definecolor{darkgreen}{rgb}{0.0, 0.5, 0.0}
\resizebox{\textwidth}{!}{%
\begin{tabular}{l|c|cc|cc|cc}
\hline
\textbf{Models} & \textbf{B} & \textbf{M1 (LLM)} & \textbf{M1} & \textbf{M2 (LLM + k)} & \textbf{M2} & \textbf{M3 (LLM + k,m)} & \textbf{M3} \\
\hline
Alibaba-NLP\_gte & 54.47 & gemma & 67.19{\color{darkgreen}$^{+12.72}$} & gemma (k=3) & 70.47{\color{darkgreen}$^{+16.00}$} & gemma (k=2,m=2) & 71.59{\color{darkgreen}$^{+17.12}$} \\
BAAI\_bge-m3 & 58.51 & glm-4 & 66.57{\color{darkgreen}$^{+8.06}$} & glm-4 (k=4) & 66.06{\color{darkgreen}$^{+7.55}$} & mistral (k=4,m=2) & 67.58{\color{darkgreen}$^{+9.07}$} \\
KaLM-Embedding-v2.5 & 77.30 & qwen & 80.06{\color{darkgreen}$^{+2.76}$} & mistral (k=2) & 81.11{\color{darkgreen}$^{+3.81}$} & mistral (k=3,m=2) & 81.18{\color{darkgreen}$^{+3.88}$} \\
Solon-embeddings-large-0.1 & 55.99 & qwen & 62.56{\color{darkgreen}$^{+6.57}$} & qwen (k=3) & 64.06{\color {darkgreen}$^{+8.07}$} & gemma (k=3,m=2) & 66.90{\color{darkgreen}$^{+10.91}$} \\
Qwen3-Embedding-0.6B & 73.54 & gemma & 76.41{\color{darkgreen}$^{+2.87}$} & gemma (k=5) & 77.34{\color{darkgreen}$^{+3.80}$} & mistral (k=5,m=4) & 77.68{\color{darkgreen}$^{+4.14}$} \\
Snowflake-arctic-embed-v2.0 & 57.21 & qwen & 65.22{\color{darkgreen}$^{+8.01}$} & mistral (k=3) & 66.44{\color{darkgreen}$^{+9.23}$} & mistral (k=5,m=1) & 68.19{\color{darkgreen}$^{+10.98}$} \\
Embeddinggemma-300m & 71.09 & glm-4 & 74.35{\color{darkgreen}$^{+3.26}$} & gemma (k=3) & 73.72{\color{darkgreen}$^{+2.63}$} & qwen (k=4,m=5) & 74.97{\color{darkgreen}$^{+3.88}$} \\
Granite-embedding-278m & 55.38 & qwen & 63.69{\color{darkgreen}$^{+8.31}$} & glm-4 (k=4) & 67.53{\color{darkgreen}$^{+12.15}$} & glm-4 (k=3,m=1) & 66.68{\color{darkgreen}$^{+11.30}$} \\
Multilingual-e5-large & 49.30 & qwen & 66.62{\color{darkgreen}$^{+17.32}$} & gemma (k=3) & 69.65{\color{darkgreen}$^{+20.35}$} & glm-4 (k=3,m=1) & 69.20{\color{darkgreen}$^{+19.90}$} \\
Modernbert-embed-large & 72.36 & gemma & 76.28{\color{darkgreen}$^{+3.92}$} & gemma (k=3) & 75.94{\color{darkgreen}$^{+3.58}$} & mistral (k=5,m=2) & 77.01{\color{darkgreen}$^{+4.65}$} \\
Mxbai-embed-large-v1 & 69.76 & qwen & 75.30{\color{darkgreen}$^{+5.54}$} & mistral (k=3) & 75.35{\color{darkgreen}$^{+5.59}$} & mistral (k=2,m=1) & 76.57{\color{darkgreen}$^{+6.81}$} \\
Llama-nemotron-embed-1b-v2 & 47.39 & gemma & 60.52{\color{darkgreen}$^{+13.13}$} & glm-4 (k=4) & 61.52{\color{darkgreen}$^{+14.13}$} & gemma (k=3,m=2) & 63.17{\color{darkgreen}$^{+15.78}$} \\
Voyage-4-nano & 75.24 & gemma & 80.90{\color{darkgreen}$^{+5.66}$} & gemma (k=1) & 80.29{\color{darkgreen}$^{+5.05}$} & mistral (k=3,m=2) & \textbf{81.42}{\color{darkgreen}$^{+6.18}$} \\
\hline
\end{tabular}%
}
\caption{F1 scores of embedding models on the N24News test set. We compare the baseline (B) with methods M1, M2, and M3. Superscripts denote the absolute improvement over the baseline ($\Delta = \text{M} - \text{B}$), and \textcolor{darkgreen}{green} indicates a positive gain.}
\label{tab:n24news_results}
\end{table*}

\subsection{Large Language Models}
We leverage four instruction-tuned large language models spanning diverse architectures
and training paradigms. The first two are dense transformer-based models:
Gemma3-27b \cite{gemmateam2025gemma3} (27B), developed by Google DeepMind with a
128K context window and distillation-based training, and Mistral-Small-3.2-24B
\cite{mistralai2025small32} (24B), a latency-optimized model from Mistral AI with
strong multilingual instruction-following and function-calling capabilities. The
remaining two feature hybrid and sparse architectures. GLM-4.5-Flash
\cite{glmteam2025glm45} (355B total, 32B active) is an open-source
Mixture-of-Experts model from Z.ai with hybrid thinking and direct-response modes
optimized for agentic and reasoning tasks. Qwen3-VL-32B-Instruct \cite{bai2025qwen3vl}
(32B) is a vision-language model from the Qwen team supporting a 256K context window
with spatial-temporal modeling via interleaved-MRoPE and DeepStack integration.

\subsection{Evaluation Protocol}

We evaluate classification performance using \textbf{Macro F1}, which equally weights all categories and is therefore appropriate for multi-class evaluation where performance across classes is equally important. For iterative refinement methods (M1--M3), the process is repeated for a maximum of $T_{\max}$ iterations, where updated definitions are proposed by the LLM using feedback from previous classification results. Confusion-aware strategies sample examples from the top-$k$ most confused class pairs, where $k \in \{1,2,3,4,5\}$. History-aware refinement additionally considers the $m$ most recent definition sets with $m \in \{1,2,3,4,5\}$. For stochastic acceptance in M3, we use a simulated annealing schedule with $T_0 = 0.1$ and $T_{\min} = 0.01$. All reported results correspond to the best-performing definitions observed during the refinement process.

\subsection{Implementation Details}

All experiments are conducted on a system equipped with 8 NVIDIA RTX 6000 Ada GPUs with 48GB VRAM each. Embedding models are evaluated using the HuggingFace Transformers framework, while large language models are served locally through the Ollama runtime. To encourage diverse candidate definitions during refinement, the temperature for LLM generation is set to 1.0. Unless otherwise specified, all embedding models remain frozen and are used in inference-only mode. Random sampling for example selection and confusion-aware refinement is performed using a fixed random seed to ensure reproducibility.

\begin{figure*}[]
    \centering
    \begin{subfigure}[b]{0.40\textwidth}
        \centering
        \includegraphics[width=\textwidth]{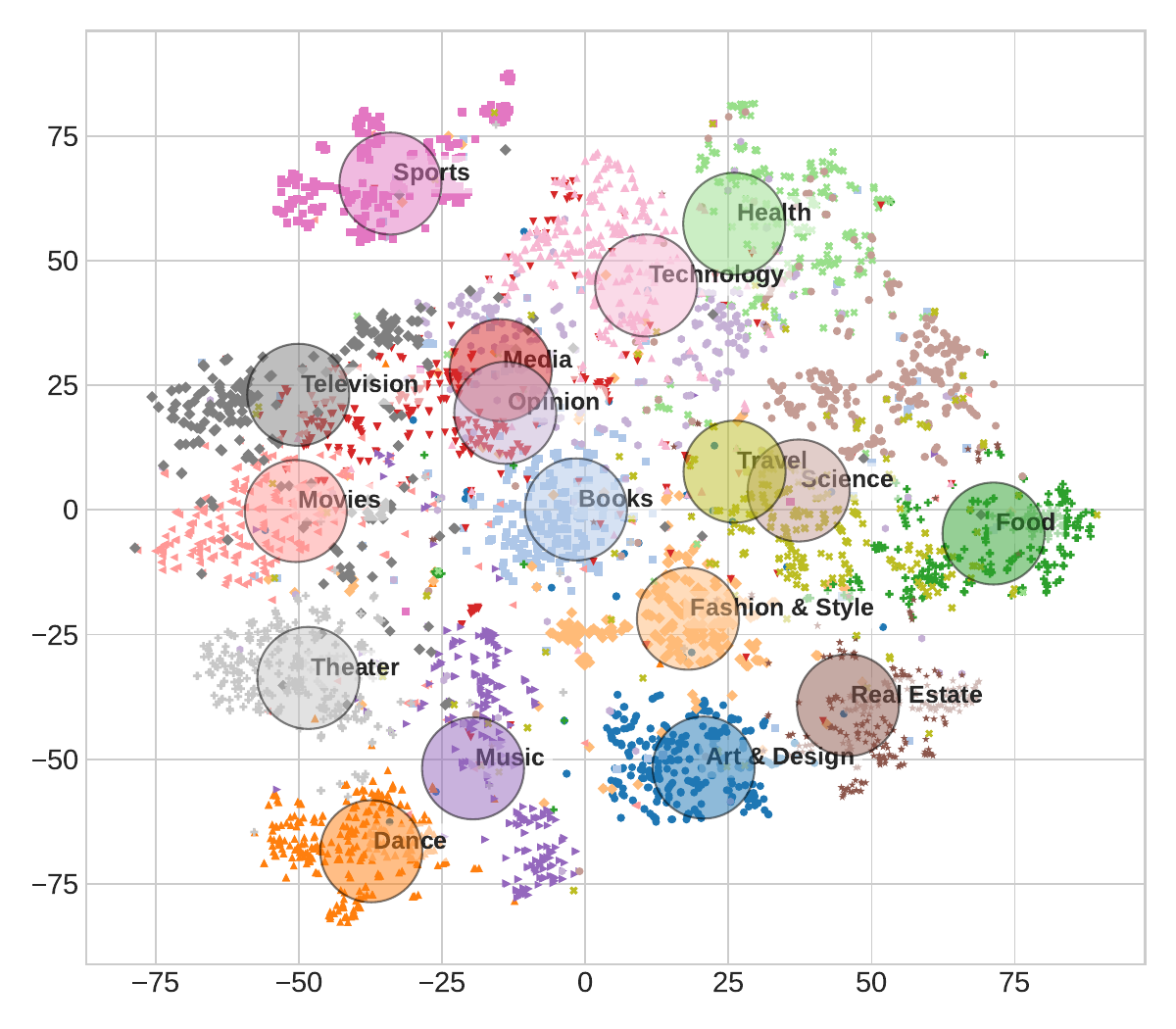}
        \caption{tSNE plot: iteration 0, F1: 71.24}
        \label{fig:tsne_iter0_n24}
    \end{subfigure}
    \begin{subfigure}[b]{0.40\textwidth}
        \centering
        \includegraphics[width=\textwidth]{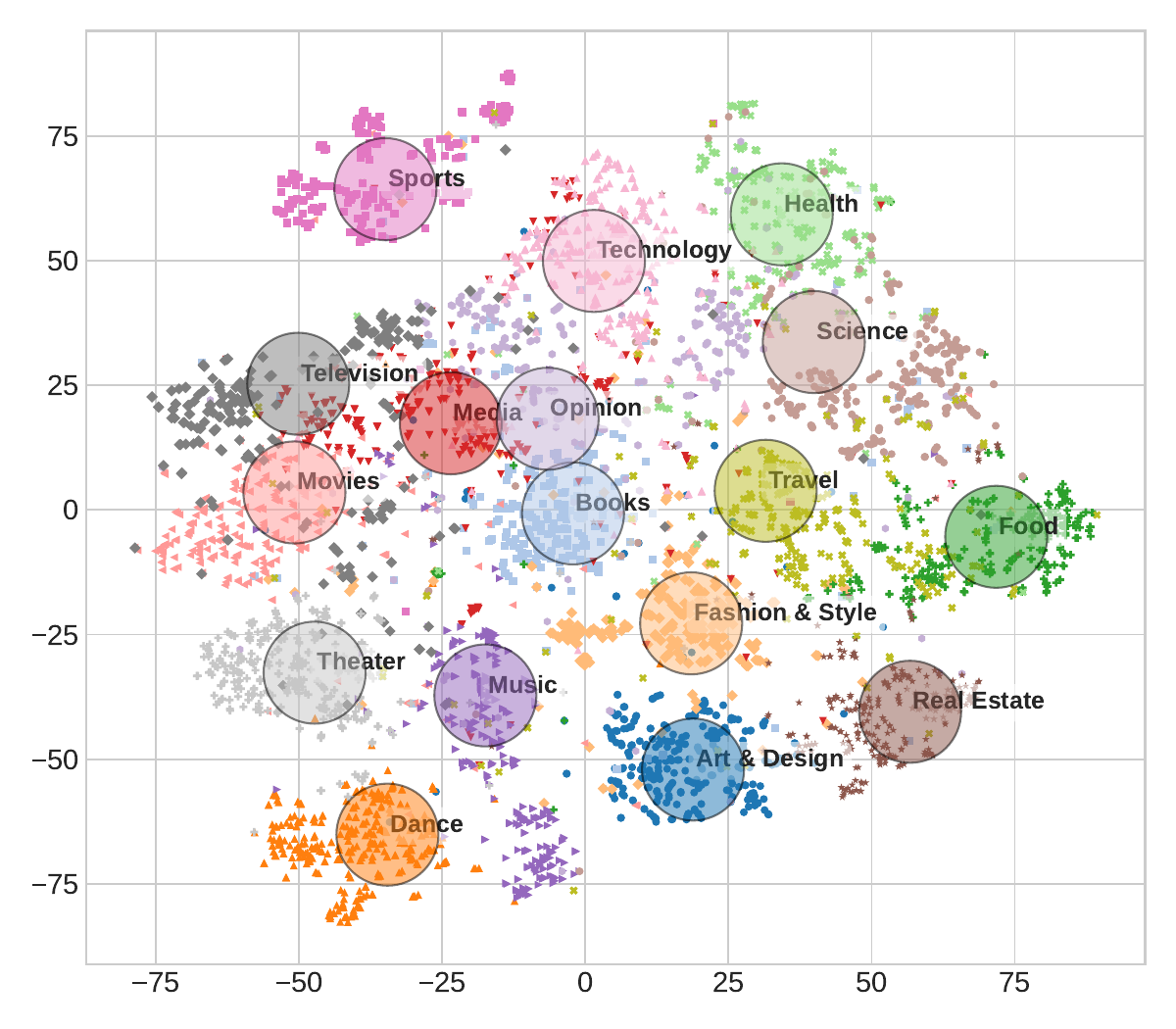}
        \caption{tSNE plot: iteration 21, F1: 81.42}
        \label{fig:tsne_iter87_n24}
    \end{subfigure}
    \caption{Disks indicate definitions. N24News (test set), Method=M3, LLM=mistral, k=3, m=2, Embed=Voyage-4-nano}
    \label{fig:tsne_n24}
\end{figure*}

\begin{figure*}[]
    \centering
    \begin{subfigure}[b]{0.40\textwidth}
        \centering
        \includegraphics[width=\textwidth]{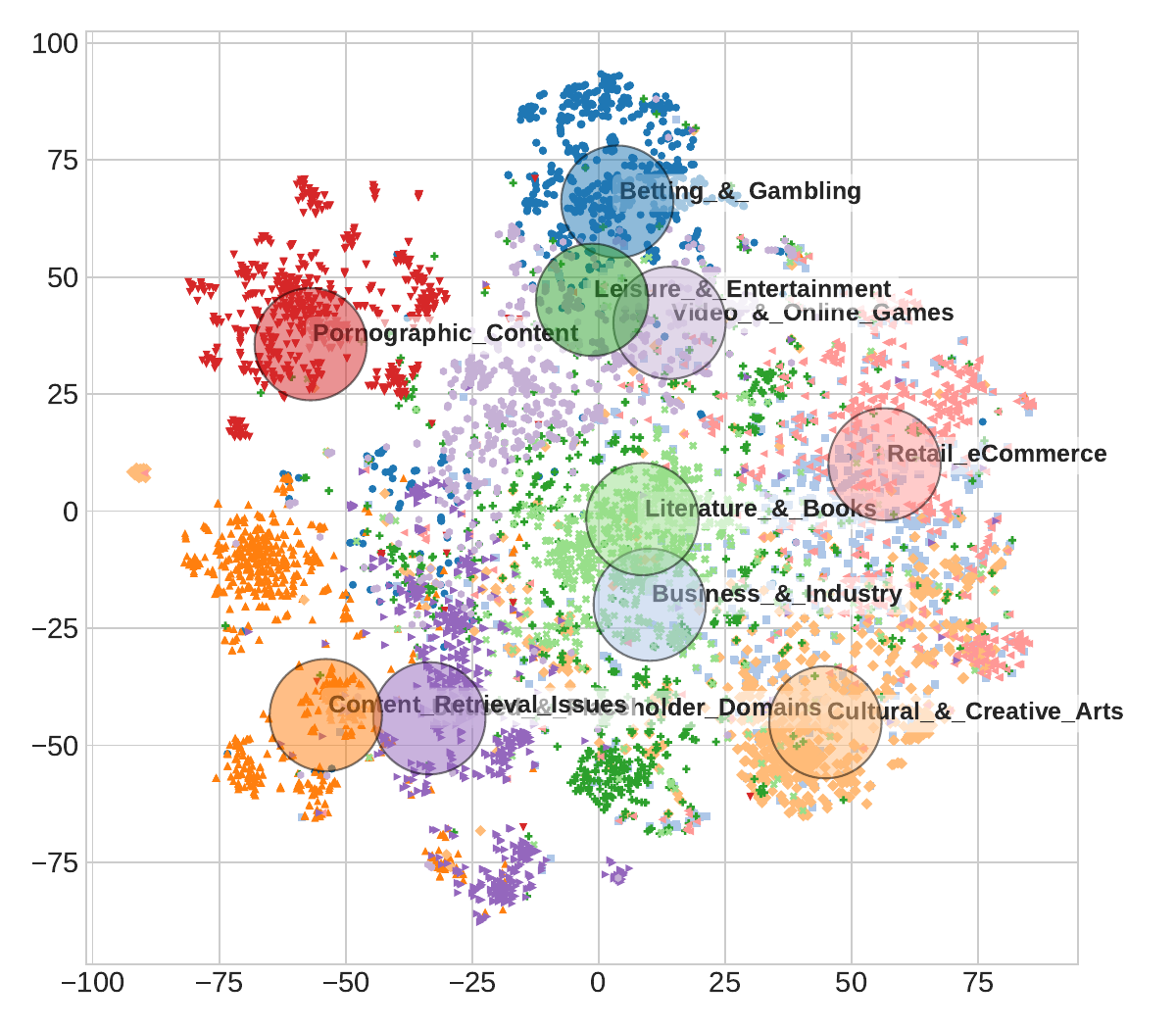}
        \caption{tSNE plot: iteration 0, F1: 62.65}
        \label{fig:tsne_b10c_iter0}
    \end{subfigure}
    \begin{subfigure}[b]{0.40\textwidth}
        \centering
        \includegraphics[width=\textwidth]{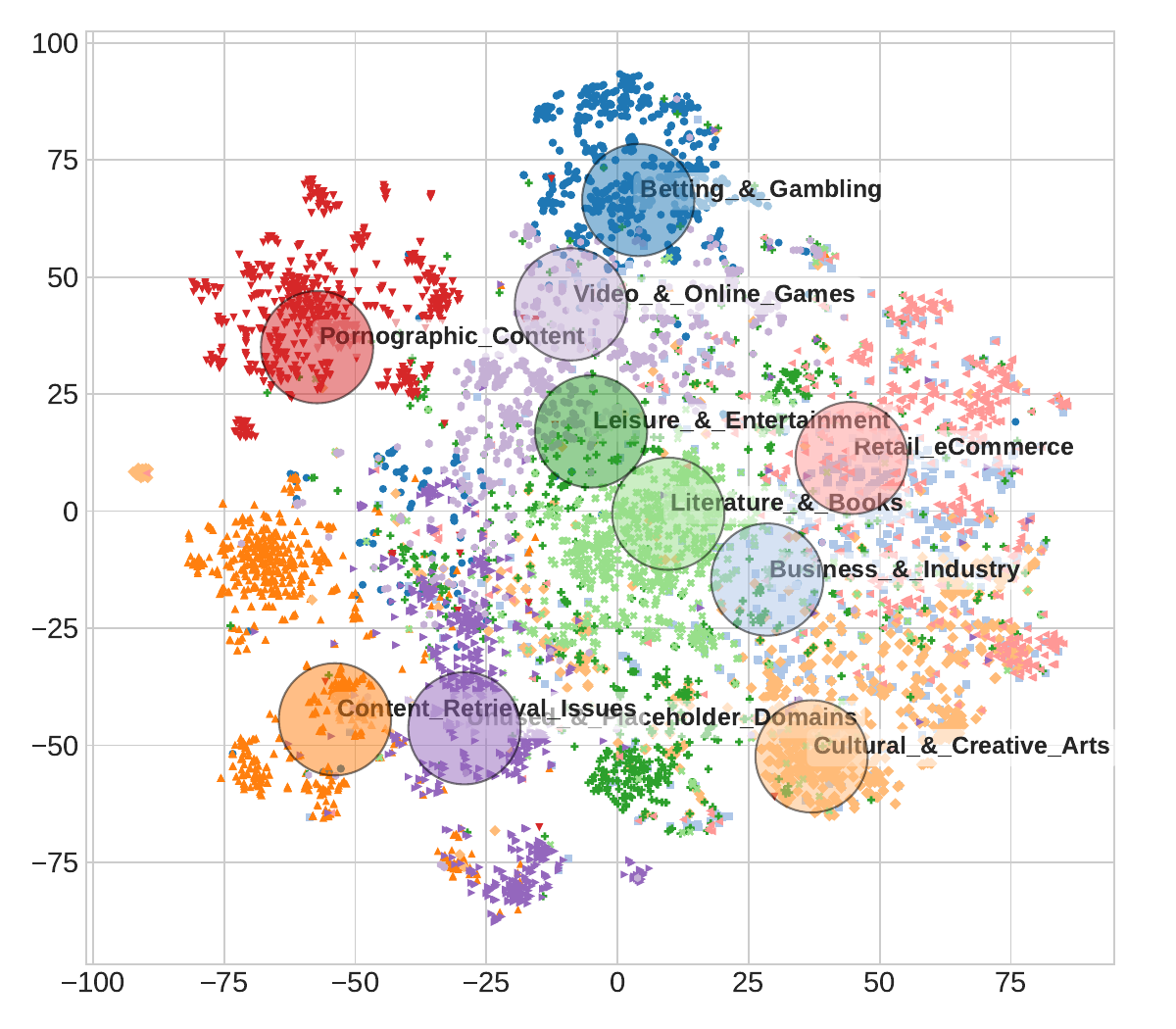}
        \caption{tSNE plot: iteration 65, F1: 75.49}
        \label{fig:tsne_b10c_iter33}
    \end{subfigure}
    \caption{Disks indicate definitions. B2MWT-10C (test set), Method=M3, LLM=mistral, k=4, m=4, Embed=Voyage-4-nano.}
    \label{fig:tsne_netstar}
\end{figure*}

\section{Results and Discussion}

\begin{figure*}[t]
    \centering
    \begin{subfigure}[b]{0.395\textwidth}
        \centering
        \includegraphics[width=\textwidth]{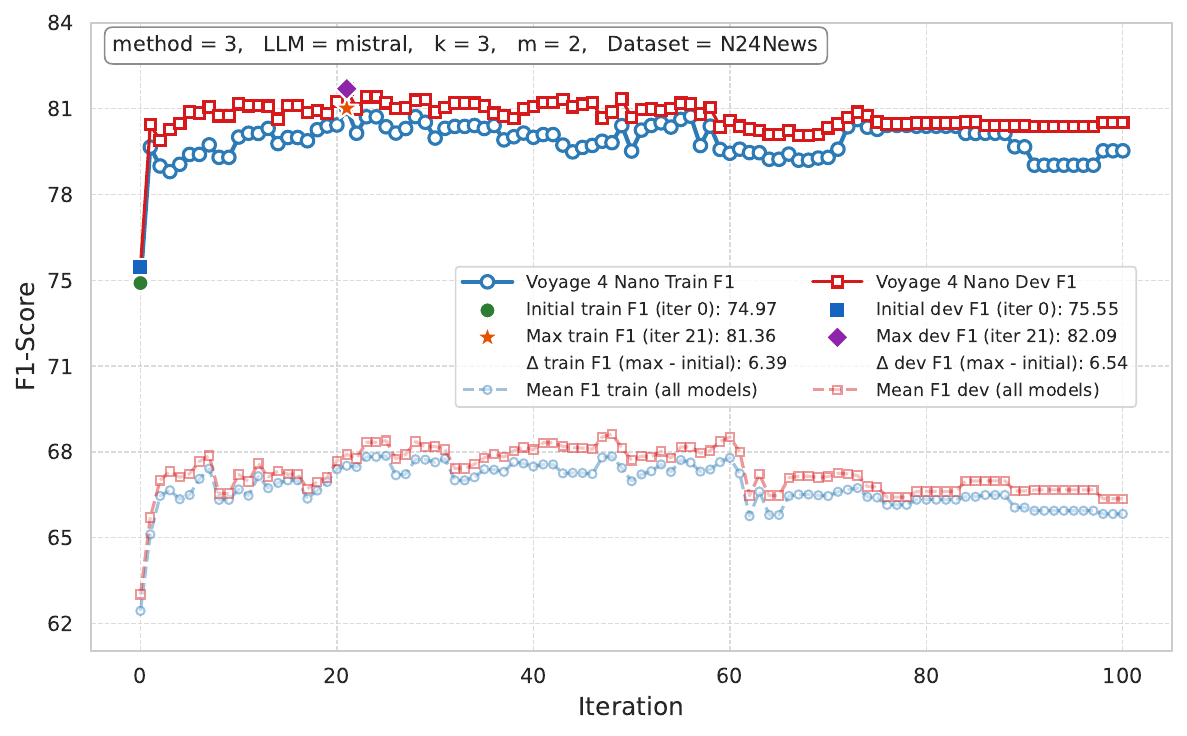}
        \caption{F1 score over N24News dataset }
        \label{fig:mean_f1_n24news}
    \end{subfigure}
    \begin{subfigure}[b]{0.395\textwidth}
        \centering
        \includegraphics[width=\textwidth]{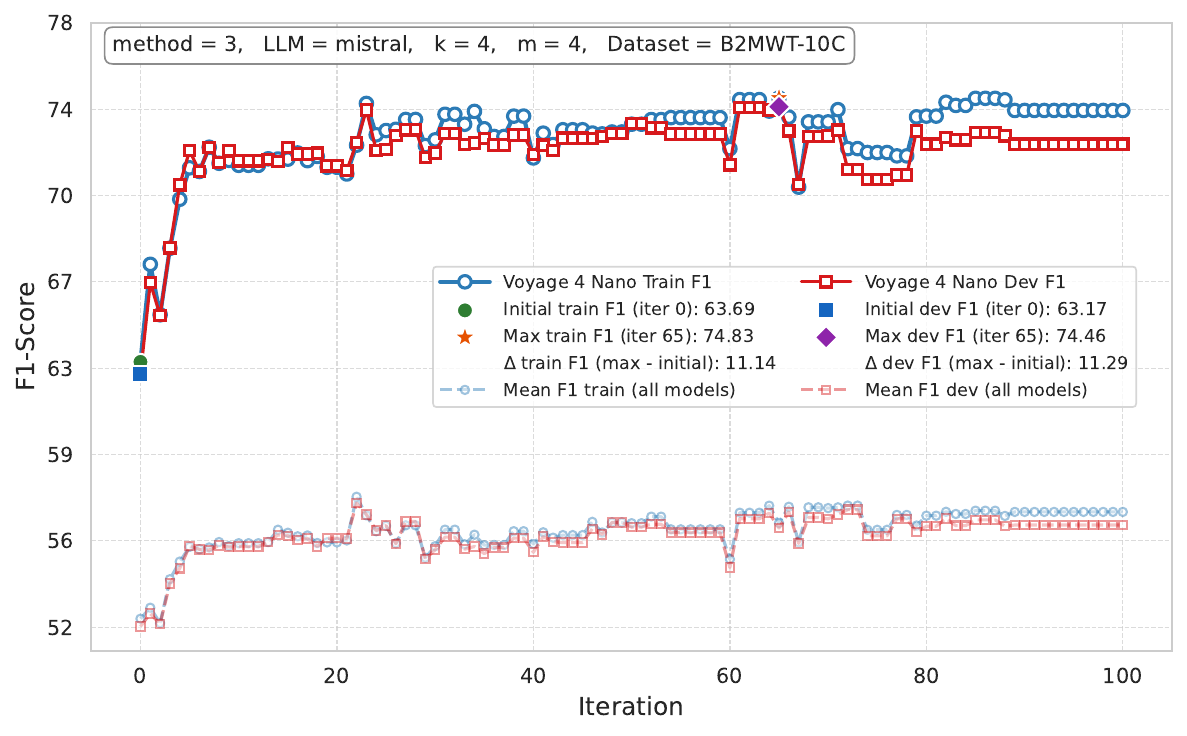}
        \caption{F1 score over B2MWT-10C dataset}
        \label{fig:mean_f1_b10}
    \end{subfigure}
    \caption{Train and dev F1 scores over refinement iterations. Dark blue and dark red lines show the train and dev F1 scores for Voyage-4-Nano, respectively; light blue and light red lines show the corresponding mean F1 scores averaged across all 13 embedding models.}
    \label{fig:f1_iter}
\end{figure*}

\begin{figure*}[t]
    \centering
    \begin{subfigure}[b]{0.39\textwidth}
        \centering
        \includegraphics[width=\textwidth]{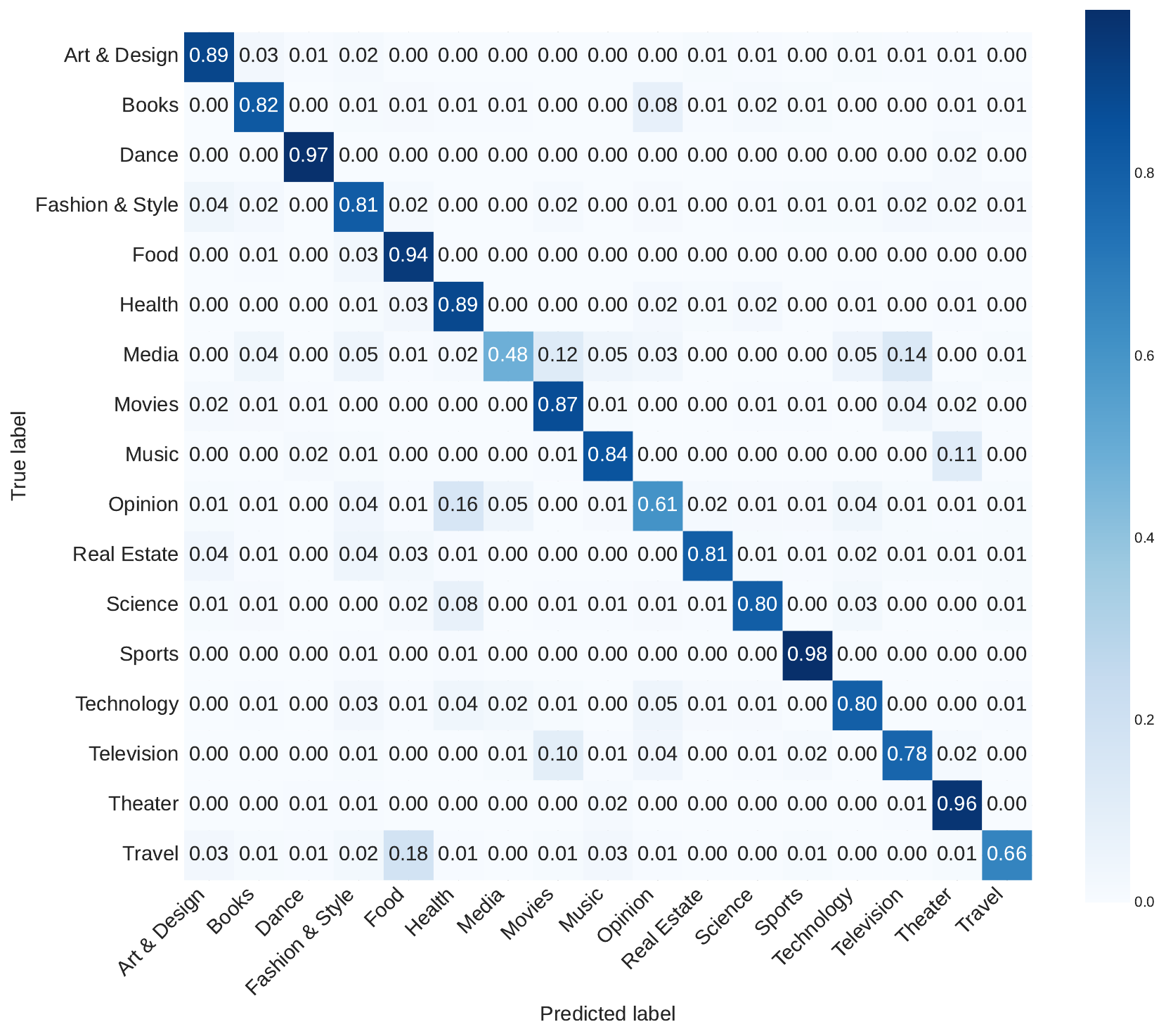}
        \caption{N24News}
        \label{fig:cm_n24news}
    \end{subfigure}
    \begin{subfigure}[b]{0.39\textwidth}
        \centering
        \includegraphics[width=\textwidth]{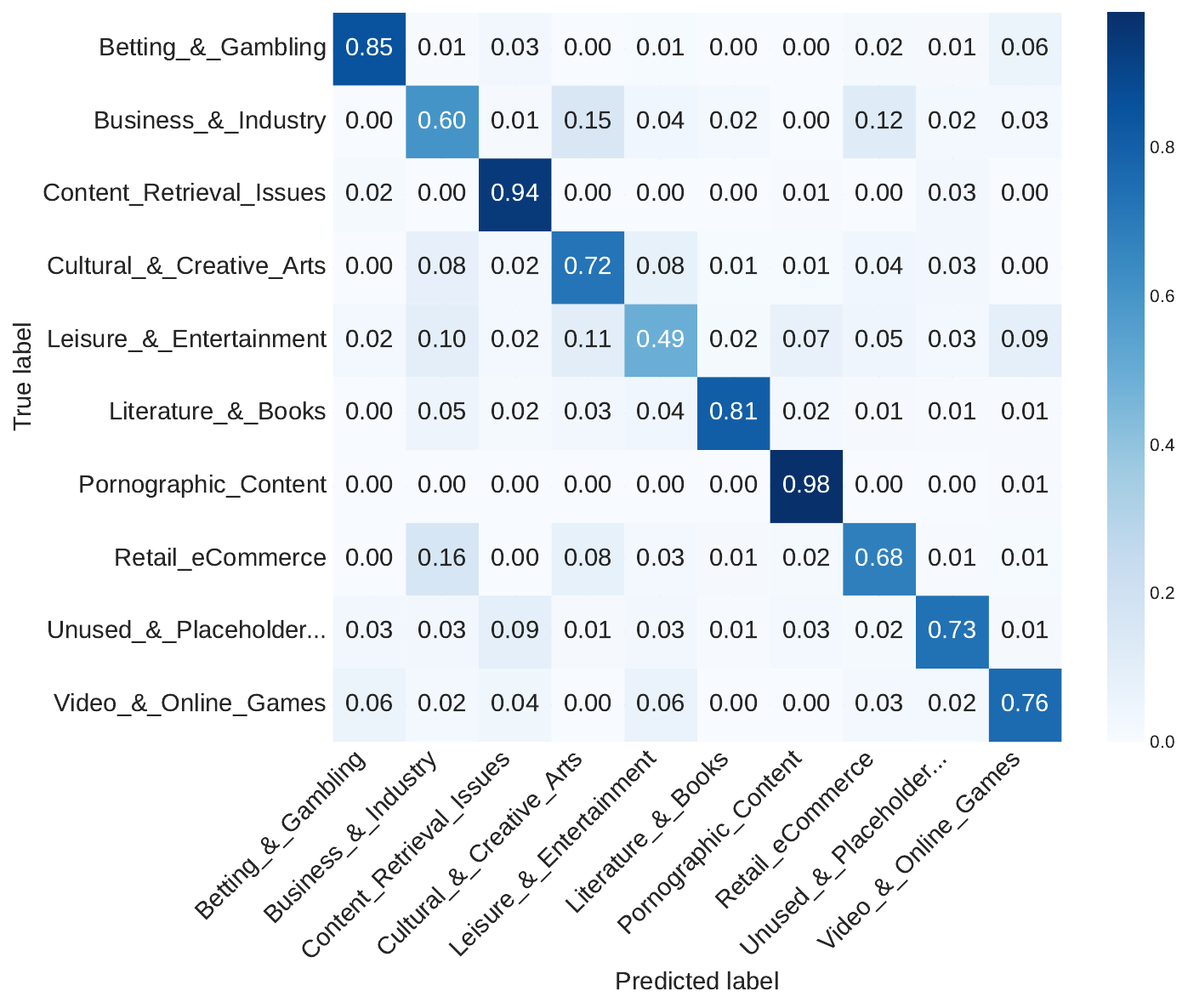}
        \caption{B2MWT-10C}
        \label{fig:cm_b10}
    \end{subfigure}
    \caption{Confusion matrix for zero-shot classification using M3 with Voyage-4-Nano model embeddings on test set.}
    \label{fig:cm_dataset}
\end{figure*}

We report Macro F1 (\%) on the B2MWT-10C and N24News benchmarks for 13 embedding models under the baseline (B) and after iterative refinement with Methods M1 (example-guided), M2 (confusion-aware), and M3 (history-aware). Tables~\ref{tab:netstar_results} and~\ref{tab:n24news_results} summarize the best configuration per method for each embedding model (LLM and hyperparameters $k$, $m$ as selected per run). Figures~\ref{fig:tsne_n24} and~\ref{fig:tsne_netstar} illustrate how definition embeddings and test points evolve in 2D (t-SNE) before and after refinement. For the N24News dataset, the "Media and Opinion" and "Travel and Science" categories initially overlapped, indicating unclear category boundaries; after the best iteration (21), these categories become well-separated. Similarly, in the B2MWT-10C dataset, the categories “Betting \& Gambling” and “Leisure \& Entertainment”, “Video \& Online Games” and “Leisure \& Entertainment”, and “Literature \& Books” and “Business \& Industry” overlap at iteration 0, but by the best iteration (65) the overlaps are resolved. Figure~\ref{fig:mean_f1_n24news} and~\ref{fig:mean_f1_b10} show the F1 score (best model) and mean F1 score (all 13 models) across refinement iterations. Both curves improve relative to the initial iteration, with larger gains in the early iterations followed by gradual convergence.  Figure~\ref{fig:cm_n24news} and~\ref{fig:cm_b10} show confusion matrices for the best M3 runs on both datasets. For the B2MWT-10C dataset, the “Business \& Industry,” “Cultural \& Creative Arts,” and “Retail \& Commerce” categories exhibit the highest confusion; details on how the definitions of these classes were updated are provided in the supplementary information (Table 4).

On the B2MWT-10C dataset (Table~\ref{tab:netstar_results}), every embedding model gains from refinement: baseline Macro F1 ranges from 41.77 (Multilingual-e5-large) to 62.65 (Voyage-4-nano), and the best M3 scores range from 49.07 to \textbf{75.49} (Voyage-4-nano, +12.84). The largest absolute gains over baseline include Embeddinggemma-300m (+19.01), Qwen3-Embedding-0.6B (+12.02), and Llama-nemotron-embed-1b-v2 (+13.49). On N24News (Table~\ref{tab:n24news_results}), baseline F1 ranges from 47.39 to 77.30; after refinement, the best M3 reaches \textbf{81.42} (Voyage-4-nano, +6.18). Multilingual-e5-large improves by +19.90 (49.30 $\rightarrow$ 69.20), and Llama-nemotron-embed-1b-v2 by +15.78 (47.39 $\rightarrow$ 63.17). In both tables, no refinement run degrades performance relative to baseline (all reported deltas are positive), and M3 often matches or exceeds the best of M1 and M2 for the same embedding model. For definition updates in M3 for both dataset, Mistral most frequently achieves the strongest performance compared to the other LLMs.

\subsection{Discussion}

The results support three main conclusions. \textbf{First}, iterative refinement of category definitions consistently improves zero-shot Macro F1 across all 13 embedding models and both datasets, without any fine-tuning of the embedding or language models. Gains are largest when the baseline is weak (e.g., Multilingual-e5-large and Llama-nemotron on N24News), suggesting that definition quality is especially critical when the embedding space alone does not separate categories well. \textbf{Second}, feeding confusion (M2) and history (M3) into the LLM yields stronger gains than example-guided refinement alone (M1). Targeting the most confused class pairs and, where applicable, the trajectory of past definitions allows the LLM to sharpen semantic boundaries and avoid repeating unhelpful edits. The strategy comparison (Table ~\ref{tab:netstar_results} and ~\ref{tab:n24news_results} ) shows that a single iteration of confusion-aware or history-aware refinement can add several points of Macro F1 over the LLM-initial baseline. \textbf{Third}, the best LLM and hyperparameters ($k$, $m$) vary by embedding model and dataset; no single configuration dominates. The t-SNE plots (Figures~\ref{fig:tsne_iter87_n24} and~\ref{fig:tsne_b10c_iter33}) illustrate that after refinement, definition embeddings tend to sit more clearly amid the test points of their respective categories, with reduced overlap between confused classes. The mean F1 curves (Figures~\ref{fig:mean_f1_n24news} and \ref{fig:mean_f1_b10}), computed on dev set, show that performance typically improves over iterations before stabilizing, consistent with the Metropolis-style acceptance rule allowing early exploration and later convergence. Confusion matrices (Figure~\ref{fig:cm_dataset}) illustrate strong diagonal dominance with minor residual confusion concentrated among semantically related categories such as media and entertainment.

\section{Conclusion}
\label{sec:conclusion}

We proposed an iterative definition refinement framework for zero-shot web content classification that improves performance by optimizing category definitions with LLMs, rather than training or fine-tuning the embedding model. Using a new 10-class, 10,000-sample human-labeled benchmark across 13 embedding foundation models, we showed that example-guided, confusion-aware, and history-aware refinement yield substantial Macro F1 gains, establishing definition quality as a critical and under-explored factor in embedding-based classification. In future work, we will study hyperparameter sensitivity and extend the framework to visual modalities and vision-language models (VLMs).

\section*{Acknowledgments}
We sincerely thank Joshi Manish, Muhammad Umair, Elias Aryee, Gaurab Khadka, Bharathi K, and Abd Ur Rehman for their valuable contributions to dataset annotation. Their dedicated efforts in labeling 10,000 samples across 10 classes (1,000 samples per class) greatly supported this work.

\small
\bibliographystyle{ieeenat_fullname}

\onecolumn
\maketitlesupplementary
\vspace{1em}

{\small
\setlength{\tabcolsep}{4.5pt}
\renewcommand{\arraystretch}{1.3}
\begin{tabularx}{\textwidth}{|c|X|X|X|}
\hline
\textbf{Iter} & \textbf{Business \& Industry} & \textbf{Cultural \& Creative Arts} & \textbf{Retail eCommerce} \\
\hline
0 &
Websites related to professional services, corporate information, industry resources, or B2B operations. They may provide company profiles, product or service offerings, industry news, or tools for business operations, excluding lifestyle services that are primarily recreational. &
Platforms showcasing artistic expression and cultural content, including design, fashion, visual arts, and creative industries. These sites often feature artist portfolios, retail of art and crafts, educational resources, related events, and announcements about new products or services. &
Websites or platforms that sell physical or digital products directly to consumers or businesses online, including lifestyle items and jewelry. These include online stores, marketplaces, and services that facilitate product browsing, purchasing, payment, and delivery, often featuring promotional messages or announcements about new products or services. \\
\hline
65 &
Websites focused on B2B transactions, industrial products, technical specifications, and professional services, often including branding, design consultancy, and custom project services for business audiences. &
Websites showcasing artistic works, creative expressions, or cultural projects such as art prints, illustrations, or artistic services, emphasizing community engagement and cultural significance, including artist profiles, galleries, and artistic descriptions without direct commercial transactions. &
Websites offering products or services for sale with clear listings, pricing, shopping cart functionality, transactional features, product filtering, promotional offers, and customer account management. \\
\hline
95 &
Websites offering technical services, industrial products, or B2B solutions tailored to industry professionals, emphasizing professional transactions, manufacturing, or infrastructure services. Excludes general consumer products, entertainment content, or personal service offerings. &
Websites showcasing original artistic works, creative services, or cultural expressions, emphasizing visual art, body art, or design projects, often with or without direct transactional features, including galleries and archives of artistic creations. Excludes commercial or entertainment content. &
Websites facilitating direct sales of consumer products or services, featuring product listings with pricing, shopping cart functionality, and transactional features. Excludes artisanal or custom product offerings tailored to industry professionals. \\
\hline
\end{tabularx}
\makebox[\textwidth][c]{%
  \begin{minipage}{\textwidth}
  \captionof{table}{Definition evolution for the three most confused B2MWT-10C classes under M3 (Voyage-4-Nano, Mistral, $k{=}5$, $m{=}1$). Initial definitions (iter.\ 0) share overlapping language around products and services. Refinement progressively introduces distinguishing criteria; B2B focus, exclusion of direct transactions, and explicit transactional features; sharpening inter-class boundaries and improving Macro F1 from 62.65\% to 75.49\%.}
  \label{tab:website_categories}
  \end{minipage}
}
}

\end{document}